\documentclass{article}
\usepackage{arxiv}
\usepackage[T1]{fontenc}
\usepackage{graphicx}
\usepackage{subcaption}
\usepackage{booktabs}
\usepackage{wrapfig}
\usepackage[table]{xcolor}
\definecolor{rowgray}{gray}{0.85}
\definecolor{rowblue}{RGB}{218,232,252}
\usepackage{multirow}
\usepackage{adjustbox}

\usepackage[colorlinks,citecolor=blue]{hyperref}
\usepackage[table]{xcolor} 
\usepackage{booktabs}

\usepackage{orcidlink}
\usepackage{amsmath}
\usepackage{amssymb}
\usepackage{cleveref}

\begin{document}

\title{Learnable Token Sparsification for Efficient Gigapixel Whole Slide Image Reasoning}

\author{
  Jingzhi Chen \\
  Shenzhen University of Advanced Technology \\
\And
  Landi He \\
   Shenzhen University of Advanced Technology \\
  \And
  Zhuo Chen \\
 Shenzhen University of Advanced Technology \\
  \And
  Shawn Young \\
 Shenzhen University of Advanced Technology \\
  \And
  Lijian Xu \thanks{Corresponding author.} \\
  Shenzhen University of Advanced Technology \\
}
\maketitle

\begin{abstract}
The processing of gigapixel whole slide images within vision language models faces a major difficulty due to an excessive number of visual tokens. Existing solutions typically rely on spatial downsampling or heuristic pruning strategies that operate without training, and these methods often discard subtle but clinically meaningful patterns because pathological evidence is scattered irregularly across the tissue. To overcome this limitation, we reformulate token reduction in whole slide images as a trainable sparsification problem, allowing the model to learn an optimal selection strategy instead of following fixed heuristics. We propose a decoupled routing architecture. To enable gradient propagation through the nondifferentiable pruning operation during training, we introduce a component called SparseLearn. This component uses a variance preserving noise gate that regulates the information flow of each patch via a differentiable Soft Top K operator, together with a diagonal attention denoiser that recovers perturbed representations without leaking spatial information. At inference time, the SparseLearn module is entirely discarded, and the trained scorer applies a deterministic Hard Top K operator to keep only the highest scoring 32 tokens, incurring no extra computation. By compressing the visual sequence down to a sparse set of just 32 tokens, which represents as little as 0.78 \% of the original length, our framework achieves 73.32 \% overall accuracy on SlideBench (TCGA), consistently surpassing sampling based baselines and general purpose vision language models. It also demonstrates strong zero shot generalization on SlideBench (BCNB) and WSI VQA*. By resolving the visual context bottleneck and preventing the dilution of sparse diagnostic evidence, this work provides a highly efficient paradigm for end to end gigapixel whole slide image reasoning.
\end{abstract}


\section{Introduction}
\label{sec:intro}
Vision-language models (VLMs) \cite{young2026scalar,xu2026unified,xu2024foundation,young2026xrayclaw} have catalyzed a paradigm shift in computational pathology, enabling interactive diagnostic reasoning and slide-level question answering. However, scaling these models to gigapixel Whole-Slide Images (WSIs) presents a fundamental computational bottleneck. A single WSI typically generates over $10^5$ patches \cite{campanella2019clinical,young2026fewer}, whose end-to-end processing amplifies the quadratic self-attention complexity and memory overhead during the large language model (LLM) prefill stage, rendering dense visual ingestion intractable \cite{zheng2022graph,pinckaers2020streaming,chen2026tc}.

To mitigate this sequence explosion, prevailing paradigms resort to training-free visual token reduction. Spatial sampling strategies restrict the input to a fixed context window by uniformly or randomly discarding most patches \cite{chen2022scaling}. Training-free pruning methods rely on proxy metrics such as attention magnitude or visual similarity \cite{rao2021dynamicvit,bolya2022tome,he2026autoselect,gao2026zerosense} to filter tokens before generative reasoning. While computationally efficient, these filters treat compression merely as identifying background noise, overlooking the extreme sparsity of pathology data where decisive diagnostic evidence, such as micro-metastases, often occupies less than 1\% of the slide.

\begin{figure}[t]
\centering
\includegraphics[width=0.999\linewidth]{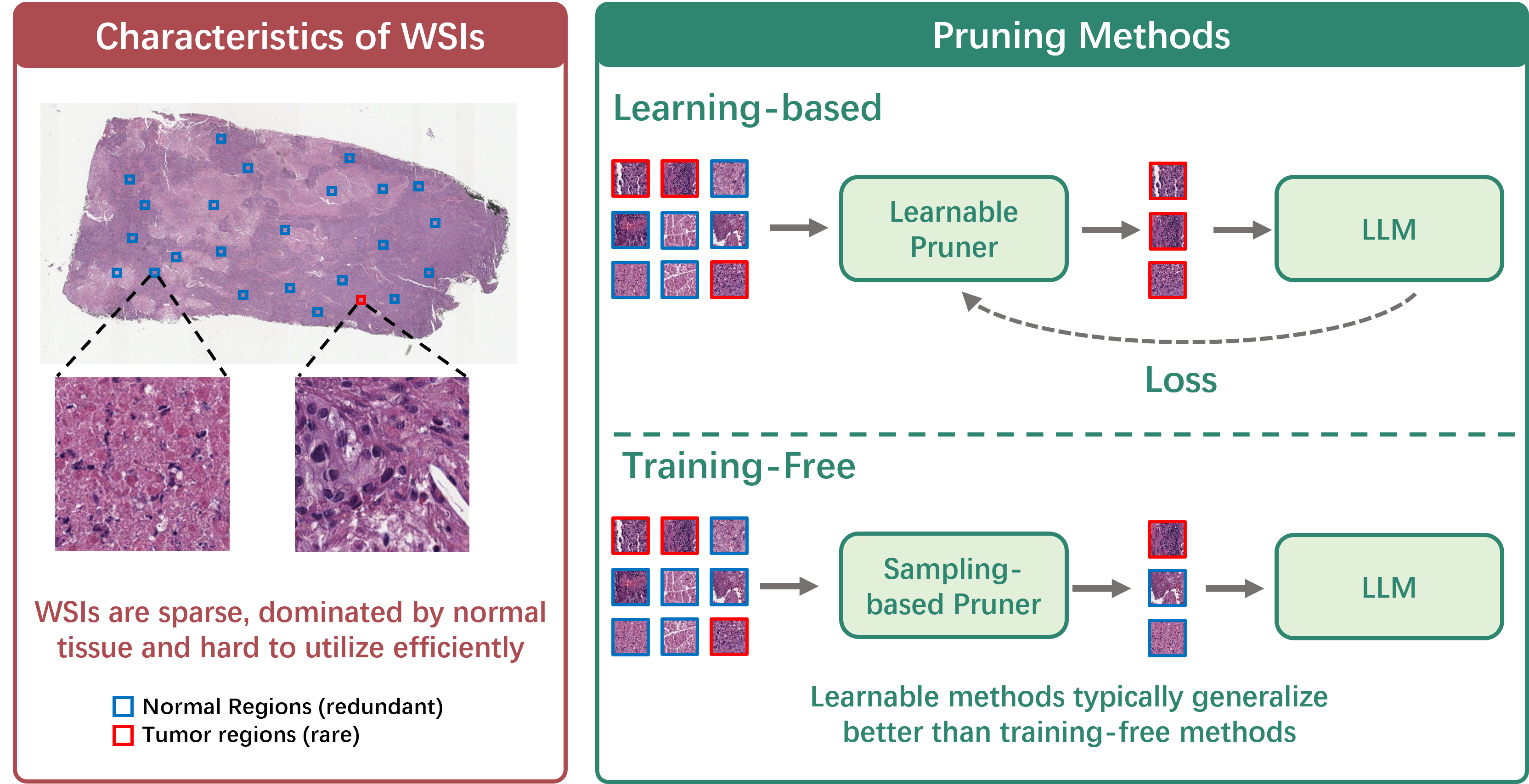}
\caption{\textbf{Illustration of WSI sparsity and pruning methods.} The left panel highlights the inherent sparsity of WSIs, where critical tumor regions are submerged in redundant normal tissue. The right panel contrasts training-free sampling with learning-based pruning, noting that the latter leverages loss feedback for better generalization in LLMs.}
\label{fig:illustration}
\end{figure}

To prevent information loss, recent efforts have shifted toward learnable token selection, exploring semantic slot aggregation for compact visual representations. Yet, routing only the most informative tokens to the LLM inherently requires discrete, hard-gating decisions such as Hard Top-$K$ \cite{kool2019stochastic}, which are fundamentally non-differentiable \cite{bengio2013estimating}. Existing frameworks circumvent this by relying on surrogate objectives, external bounding-box annotations, or intrusive routing layers deep within transformer blocks \cite{takezoe2026learnpruner}, disrupting pre-trained language priors and hindering end-to-end multimodal alignment (see \Cref{fig:illustration}).

We address this bottleneck by reconceptualizing WSI token pruning as \textit{learnable sparsity}. Instead of static, training-free spatial discard, we enable the model to autonomously learn an optimal token pruning strategy via continuous optimization. We propose a decoupled routing framework. To establish continuous gradients for the non-differentiable pruning operation during training, we introduce SparseLearn. Guided by a Soft Top-$K$ operator, SparseLearn employs a continuous variance-preserving (VP) noise gate to modulate each patch's information flow, creating continuous gradients without removing tokens. A diagonal-attention Denoiser subsequently repairs latent distribution shifts while strictly preventing spatial information leakage. At inference, the SparseLearn module is entirely detached; a deterministic Hard Top-$K$ operator executes token pruning based on the trained Scorer, adding zero computational latency.

Our contributions are summarized as follows:

(1) We reformulate WSI token pruning from training-free pruning into a learnable sparsity paradigm, optimized end-to-end through continuous gradients without auxiliary losses or architectural intrusions into the frozen LLM.

(2) We introduce a decoupled routing mechanism. To bridge the optimization gap of discrete Top-$K$ token selection, we design SparseLearn, a training module pairing a VP noise gate with diagonal-attention denoising to establish continuous gradients.

(3) Empirical evaluations validate our framework under a strict sparsity budget. Enforcing a fixed 32-token budget, our approach achieves an average 58-fold reduction in visual sequence length and a maximum 128-fold dynamic compression limit. Under this compact regime, our framework yields 73.32\% overall accuracy on SlideBench (TCGA), with competitive zero-shot generalization on SlideBench (BCNB) and WSI-VQA*.

\section{Related Work}
\label{sec:related-work}

\subsection{WSI Representation Paradigm}

Gigapixel WSIs analysis is predominantly formulated as weakly supervised multiple instance learning. Early frameworks like AB-MIL \cite{ilse2018attention} and CLAM \cite{lu2021data} introduced attention-based aggregation but overlooked spatial correlations. Transformer-based architectures such as TransMIL \cite{shao2021transmil} subsequently captured tissue-wide dependencies, with recent advances incorporating mixture-of-experts \cite{hashimoto2024multimodal,wu2025learning}, dynamic fusion \cite{cao2023multi}, and slide-level supervision \cite{tang2024feature,tang2026revisiting}. Pathology-specific foundation models \cite{chen2024towards,xu2024whole,wu2026multimodal,xu2024medvilam}, including vision-language \cite{huang2023visual,lu2024visual} and multimodal variants \cite{sun2025cpath}, now encode raw patches as high-dimensional morphological tokens. Other representation learning strategies \cite{yang2024segmentation,yang2023geometry,feng2026efficient} have also been explored in medical imaging contexts.

Despite these advances, a standard slide yields hundreds of thousands of instances, introducing the gigapixel curse \cite{campanella2019clinical,chen2022scaling}. Global self-attention incurs $\mathcal{O}(N^2)$ complexity, which becomes critical when deploying Multimodal LLMs \cite{li2023llava,hurst2024gpt,he2024meddr,seyfioglu2024quilt} with strictly bounded context windows. Existing remedies such as aggressive downsampling, sequence truncation \cite{campanella2019clinical}, or regional slicing like HIPT \cite{chen2022scaling} mitigate the quadratic explosion but disrupt global spatial continuity.

More importantly, these strategies process the tissue without spatial awareness, failing to address the extreme heterogeneity where crucial diagnostic indicators occupy a minute fraction of the tissue. Random sampling or threshold-based filtering largely fails in this sparse regime, rendering models susceptible to discarding scarce but decisive tokens. Resolving this demands explicit optimization of token selection under a strict sparsity budget, ensuring critical morphological features are preserved without information dilution.

\subsection{Visual Token Redundancy and Reduction}

A natural approach to the sequence length crisis is to adopt token reduction strategies from general computer vision, which fall into trainable pruning \cite{wu2026hidrop,yang2025one,he2026beyond} and training-free methods \cite{bolya2022tome,dong2025mmtok,zhang2026beyond}. However, directly transplanting these algorithms to gigapixel WSI inference exposes fundamental limitations.

Training-free methods, while computationally efficient, are sensitive to attention biases such as attention sinks and dispersion \cite{yang2025visionzip}, often retaining semantically uninformative tokens under high compression ratios and degrading noticeably under aggressive pruning \cite{shao2025tokens}. Their generalization also remains limited, as they rely heavily on model-specific statistics. Trainable pruning frameworks attempt to learn dynamic token routing but face a deeper bottleneck: hard pruning severs gradient flow, necessitating gradient approximations like the Straight-Through Estimator (STE) \cite{bengio2013estimating,takezoe2026learnpruner} or Gumbel-Softmax relaxation\cite{jang2016categorical}. While STE-induced bias is tolerable in densely supervised natural images, it becomes catastrophic under the sparse, weakly supervised gigapixel WSIs setting where the approximated gradients fail to credit scarce tumor tokens, causing the model to collapse into degenerate minima and discard critical signals as background.

To address these challenges, emerging efforts explore specialized token pruning paradigms \cite{hu2025loc,wang2026wsisum,guo2025focus}. Yet a critical gap remains: the field lacks a mathematically rigorous mechanism that avoids both the semantic dilution of training-free blending and the optimization collapse of STE-based routing. Diagnostic token retention must instead be governed by stable, continuous gradient flow under a fixed sparsity budget, ensuring robust convergence without sacrificing scarce pathological signals.

\section{Methodology}
\label{sec:methodology}

\begin{figure}[t]
\centering
\includegraphics[width=0.999\linewidth]{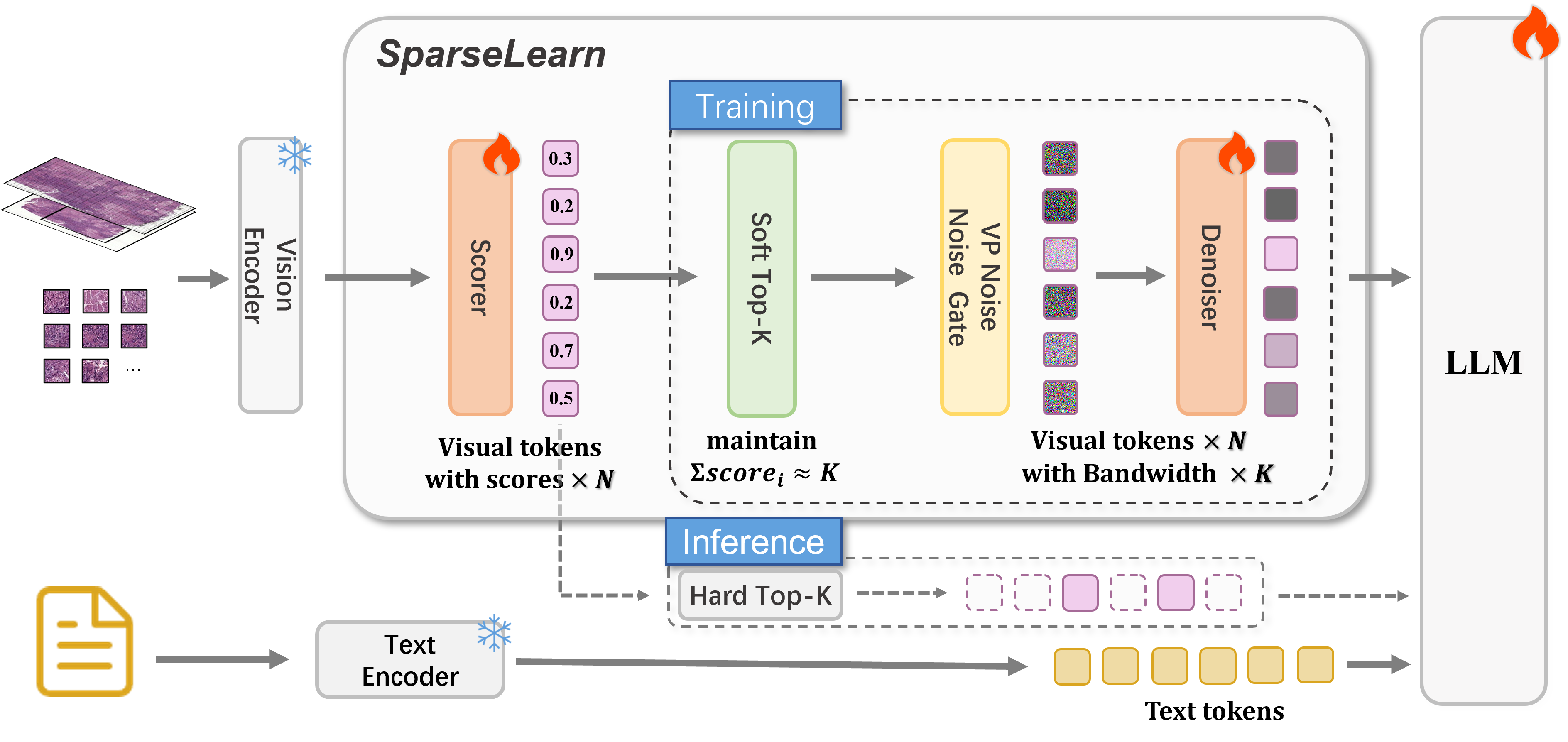}
\caption{\textbf{Overview of our framework.} Frozen encoders extract visual and textual features. To mitigate LLM context bottlenecks, a learnable sparsity mechanism dynamically reduces visual tokens via decoupled routing. \textbf{Training path}: To enable continuous gradients for discrete pruning, the SparseLearn module computes token importance ($s_i$) via a Scorer, mapped to continuous weights ($\alpha$) using Soft Top-$K$. A VP Noise Gate injects $\alpha$-scaled VP noise for continuous optimization without token removal, while a Denoiser recovers perturbed representations. \textbf{Inference path}: SparseLearn is detached. Guided solely by the trained Scorer, a Hard Top-$K$ operator executes deterministic token pruning, retaining critical pathology features with zero computational overhead. Fire and snowflake icons denote trainable and frozen modules.}
\label{fig:framework_patho}
\end{figure}

\subsection{System Formalization}

Given a WSI $\mathcal{I}$, we first discretize it into a sequence of $N$ non-overlapping patches $\{p_n\}_{n=1}^N$, and extract the corresponding visual feature sequence:
\begin{equation*}
    \mathbf{X}^v = \{x_1, x_2, \dots, x_N\} \in \mathbb{R}^{N \times d_v}
\end{equation*}
where $d_v$ denotes the feature dimension of the visual encoder. This step establishes the initial comprehensive representation of the visual information. 

However, the extensive visual sequence of length $N$ generated by WSI imposes a quadratic computational and memory burden on the self-attention mechanism. Given that the forward pass of the visual encoder accounts for less than 5\% of the total overhead, whereas the LLM's prefill stage for processing long visual prefixes constitutes the primary computational bottleneck, we introduce a learnable token pruning framework between the visual encoder $\mathcal{E}_v$ and the projection layer $\mathbf{P}_{v \to \ell}$. \Cref{fig:framework_patho} illustrates the complete pipeline, wherein WSI patches and text prompts are processed in parallel before fusion (a), and the token routing process is decoupled into two equivalent paths for training and inference (b). To drive the training phase without losing differentiability, we introduce \textbf{SparseLearn}, a specialized training module comprising a learnable scorer $\mathcal{S}_\theta$ (parameterized by $\theta$) and a lightweight denoiser $\mathcal{D}_\omega$ (parameterized by $\omega$). These components aim to learn an optimal pruning strategy under a strict sparsity budget $K$ ($K \ll N$).

\textbf{Training path.} During training, the scorer $\mathcal{S}_\theta$ computes per-token importance scores, which are subsequently polarized by a differentiable Soft Top-$K$ operator under the fixed sparsity budget $K$. Instead of directly discarding tokens—which would sever gradients—we maintain the full sequence length $N$ by applying a VP noise gating mechanism within the SparseLearn module. Noise is injected with a magnitude inversely proportional to each token's polarized score, generating a continuous noise-injected sequence $\tilde{\mathbf{X}}^v$. The lightweight denoiser $\mathcal{D}_\omega$ then processes this perturbed sequence to restore the perturbed representations before forwarding the representations through the projection layer $\mathbf{P}_{v \to \ell}$ to the LLM. We jointly optimize the scorer parameters $\theta$, the denoiser parameters $\omega$, and the low-rank adaptation (LoRA) parameters $\phi$ of the LLM via the negative log-likelihood (NLL) loss for next-token prediction:
\begin{equation}
    \min_{\theta, \omega, \phi} \mathcal{J}_{\text{NLL}} \left( f_{\text{LLM}} \left( [ \mathcal{D}_\omega(\tilde{\mathbf{X}}^v) \mathbf{P}_{v \to \ell} ; \mathcal{E}_t(T) ] ; \Theta, \phi \right), \{y_t^*\} \right)
\end{equation}
where $\Theta$ denotes the frozen original parameters of the LLM, $\mathcal{E}_t(T)$ is the text prompt embedding, and $\{y_t^*\}$ represents the ground-truth targets.

\textbf{Inference path.}
To execute the learned token pruning, the denoiser and noise injection mechanisms of SparseLearn are completely detached. The system relies entirely on the trained scorer $\mathcal{S}_\theta$ to evaluate token importance and executes a deterministic hard pruning strategy. It directly retains the top $K$ most representative feature vectors via a standard hard Top-$K$ operator:
\begin{equation}
    \hat{\mathbf{X}}^v = \text{Top-}K(\mathbf{X}^v, \mathcal{S}_\theta(\mathbf{X}^v), K) \in \mathbb{R}^{K \times d_v}
\end{equation}
In this process, the retained tokens $\hat{\mathbf{X}}^v$ must preserve their spatial position indices from the original sequence. This design ensures that the rotary position embedding (RoPE) within the LLM can accurately capture the geometric topological correlations among tokens, a property crucial for understanding the spatial morphology of pathological tissues.

Through this two-stage formalization, we reformulate WSI token pruning as a learnable sparsity problem. Under this framework, the scorer $\mathcal{S}_\theta$ is optimized under a strict sparsity budget, where it must learn to prioritize the most informative tokens. It must learn to identify and prioritize the transmission of critical pathological semantics for downstream diagnostic reasoning under a strict sparsity budget, thereby achieving efficient end-to-end analysis of high-resolution WSI.

\subsection{Learnable Token Scorer with Soft Top-\textit{K} Selection}

The scorer $\mathcal{S}_\theta$ comprises $L$ lightweight Transformer encoder blocks followed by a linear projection, mapping each of the $N$ tokens to a scalar importance score:
\begin{equation}
\mathbf{s} = \mathcal{S}_\theta(\mathbf{X}^v) \in \mathbb{R}^{N}.
\end{equation}
To transform these raw scores into continuous weights under a strict sparsity budget, we employ the Soft Top-$K$ operator~$\Phi_K$~. Raw scores are first z-score normalized for numerical stability, then mapped through~$\Phi_K$:
\begin{equation}
\boldsymbol{\alpha} = \Phi_K\!\left(\hat{\mathbf{s}} \,/\, \tau \right) \in [0,1]^{N},
\quad \text{with}\quad
\sum_{i=1}^{N} \alpha_i \approx K,
\label{eq:soft_topk_patho}
\end{equation}
where $\hat{\mathbf{s}}$ denotes the normalized scores and $\tau > 0$ is the temperature parameter. Conceptually, $\Phi_K$ functions as a smooth, temperature-scaled mapping similar to softmax. The primary difference lies in the normalization constraint: unlike softmax, which requires $\sum_i \alpha_i = 1$, $\Phi_K$ enforces $\sum_i \alpha_i = K$, modifying the operation into a sparsity-constrained soft assignment. 

Furthermore, an implicit data-dependent threshold distinguishes $\Phi_K$ from softmax by pushing the output scores toward $0$ or $1$, resulting in a bimodal distribution instead of being spread across all tokens. Given the fixed sparsity budget $K$, the scorer naturally learns to identify \emph{which} tokens to retain, rather than determining \emph{how many}. During training, $\tau$ is annealed from $\tau_{\mathrm{start}}$ to $\tau_{\mathrm{end}}$ using a cosine schedule. At larger values of $\tau$, the assigned weights remain diffuse, and as $\tau \to 0$, they asymptotically approximate the deterministic binary mask used for token pruning at inference.

\subsection{Continuous Gradients via Noise Injection}

Having obtained the polarized importance scores $\boldsymbol{\alpha}$ from the Soft Top-$K$ operator, we must restrict the information transmission to the downstream LLM. Implementing a standard hard Top-$K$ pruning directly discards low-scoring tokens, introducing a discrete discontinuity that produces zero gradients almost everywhere and fundamentally prevents end-to-end optimization. To circumvent this non-differentiability and establish continuous gradients while strictly enforcing the sparsity budget, we maintain the constant sequence length $N$ and modulate the token-wise information through continuous noise injection.

Specifically, we implement a VP noise gating mechanism. For the $i$-th visual token $x_i \in \mathbb{R}^{d_v}$, the noise-injected representation $\tilde{x}_i$ is formulated as:
\begin{equation}
    \tilde{x}_i = \sqrt{\alpha_i}\, x_i + \sqrt{1 - \alpha_i}\, \boldsymbol{\epsilon}_i, \quad \boldsymbol{\epsilon}_i \sim \mathcal{N}(\mathbf{0}, \mathbf{I}_{d_v})
    \label{eq:vp_noise_patho}
\end{equation}
where $\mathbf{I}_{d_v} \in \mathbb{R}^{d_v \times d_v}$ is the identity matrix. From an information-theoretic perspective, this operation establishes a differentiable proxy for discrete token removal. When $\alpha_i \to 1$ for diagnostically critical tokens, the noise component diminishes, preserving the original feature; when $\alpha_i \to 0$, the visual signal is subsumed by isotropic Gaussian noise. Intermediate values yield a smooth interpolation between these two states.

To ensure $\text{Var}(\tilde{x}_i) \approx \text{Var}(x_i)$, we design the noise injection using coefficients $\sqrt{\alpha_i}$ and $\sqrt{1 - \alpha_i}$. This VP formulation is mathematically grounded by the fact that standard visual encoders typically output layer-normalized features. Preserving this latent feature scale is critical to preventing distribution shifts that would otherwise destabilize the LLM. Through this explicit design, the scorer dictates the noise magnitude—and consequently the effective information flow—of each token during optimization. As empirically verified in Section 4, this continuous VP gating successfully simulates the information restriction behavior of deterministic hard Top-$K$ token pruning.

\subsection{Lightweight Denoiser with Diagonal Attention}

While the VP gating ensures macroscopic variance stability across the feature sequence, the localized addition of Gaussian noise inherently perturbs the latent distribution of individual tokens. This deviation is most severe for tokens assigned low importance scores, where the visual signal is largely overshadowed by noise. To realign these perturbed representations with the manifold expected by the frozen LLM, we append a lightweight denoiser $\mathcal{D}_\omega$ instantiated as a single Transformer encoder block.

Applying standard global self-attention over the noise-injected sequence would allow semantic details to propagate from high-fidelity tokens to heavily noised ones via the attention matrix. Such cross-token communication would inadvertently circumvent the precise sparsity bottlenecks established by the preceding VP gate. To eliminate this risk, we implement a \emph{diagonal attention} strategy: an identity mask isolates each token, restricting it to self-reference only. Consequently, the self-attention module reduces to a point-wise nonlinear mapping, operating strictly through the value projection and feed-forward networks. This structural constraint guarantees zero spatial leakage while providing the necessary representational capacity to bridge the distribution gap. The denoiser is not utilized during inference, incurring zero computational latency.

\section{Experiments}
\label{sec:experiments}

\subsection{Experimental Setup and Implementation Details}

\textbf{Datasets and Evaluation Protocol.} 
Our model is trained and evaluated on the SlideBench (TCGA) benchmark \cite{chen2025slidechat}, reporting both overall accuracy and performance stratified across two critical categories: \textbf{Microscopy} assesses the description of low-level morphological and staining features; \textbf{Diagnosis} evaluates histology-based reasoning and clinical subtyping. To assess out-of-distribution robustness, we further conduct zero-shot generalization tests on SlideBench (BCNB) and the external WSI-VQA* dataset \cite{chen2024wsi}.

\textbf{Baselines.} 
To rigorously validate our framework across diverse diagnostic dimensions, we construct a comprehensive baseline matrix. This matrix includes theoretical upper-bound uncompressed models like SlideChat \cite{chen2025slidechat}, which retain full-slide evidence without token reduction; compute-matched spatial sampling and general-purpose VLMs, including LLaVA-Med \cite{li2023llava}, Quilt-LLaVA \cite{seyfioglu2024quilt}, MedDr \cite{he2024meddr}, and the proprietary GPT-4o \cite{hurst2024gpt}, which represent current paradigms for training-free or large-scale multimodal reasoning.

\textbf{Training and Optimization.} 
The architecture is optimized utilizing two NVIDIA A6000 GPUs via a progressive two-stage training protocol. In Stage 1, the LLM backbone is strictly frozen, and optimization is restricted to the active components: the slide encoder, the SparseLearn module, and the cross-modal projector. In Stage 2, to prevent catastrophic forgetting of learned spatial priors, the slide encoder is frozen. We continue updating the SparseLearn module and the projector, while simultaneously introducing parameter-efficient fine-tuning (LoRA) to the LLM.

\begin{table}[htbp]
\centering
\caption{Main results on SlideBench (TCGA) and zero-shot generalization on SlideBench (BCNB) and WSI-VQA*.}
\label{tab:combined_results_patho}
\renewcommand{\arraystretch}{1.2} 
\setlength{\tabcolsep}{3pt} 

\definecolor{mygray}{gray}{0.9}
\definecolor{myblue}{rgb}{0.88, 0.94, 0.99}

\resizebox{\linewidth}{!}{
\begin{tabular}{lcccccc}
\toprule
\multirow{2}{*}{\textbf{Methods}} & \multirow{2}{*}{\textbf{FLOPs}} & \multicolumn{3}{c}{\textbf{SlideBench (TCGA)}} & \textbf{SlideBench} & \textbf{WSI} \\
\cmidrule(lr){3-5} 
 & & \textbf{Micros.} & \textbf{Diagnos.} & \textbf{Overall} & \textbf{(BCNB)} & \textbf{(VQA*)} \\
\midrule
\rowcolor{mygray} \multicolumn{7}{c}{(Upper Bound)} \\
\textcolor{gray}{SlideChat} & \textcolor{gray}{133.3T} & \textcolor{gray}{87.64} & \textcolor{gray}{73.27} & \textcolor{gray}{81.17} & \textcolor{gray}{54.14} & \textcolor{gray}{60.18} \\
\midrule
\rowcolor{mygray} \multicolumn{7}{c}{(token pruning $\sim60\times$)} \\
Random Baseline & -- & 24.44 & 24.91 & 25.02 & 24.40 & 24.14 \\
GPT-4o & -- & 62.89 & 46.69 & 57.91 & 41.43 & 30.41 \\
LLaVA-Med & 1.70T & 47.34 & 32.78 & 42.00 & 30.10 & 26.31 \\
Quilt-LLaVA & 1.70T & 57.76 & 35.96 & 48.07 & 32.19 & 44.43 \\
MedDr & 1.70T & 73.30 & 57.78 & 67.70 & 33.67 & 54.36 \\
\rowcolor{myblue} \textbf{Ours} & 1.73T & \textcolor{red}{\textbf{81.68}} & \textcolor{red}{\textbf{70.09}} & \textcolor{red}{\textbf{73.32}} & \textcolor{red}{\textbf{56.89}} & \textcolor{red}{\textbf{60.76}} \\
\bottomrule
\end{tabular}
} 
\end{table}

\subsection{Main Results}

\textbf{SlideBench Benchmark and Zero-Shot Evaluation.}
\Cref{tab:combined_results_patho} summarizes the quantitative results on the SlideBench (TCGA) benchmark. Under a stringent capacity constraint, our approach achieves 73.32\% overall accuracy, establishing a definitive performance margin over compute-matched spatial sampling paradigms. This gap validates that our learnable semantic selection effectively preserves sparse diagnostic evidence typically discarded by spatial sampling. Crucially, on the Diagnosis subset, our framework (70.09\%) retains over 96\% of the reasoning capacity of the uncompressed SlideChat upper bound (73.27\%) while consuming merely 1.3\% of its computational overhead. This performance extends to zero-shot settings consistently outperforming all equivalent compressed baselines and GPT-4o. This robust generalization confirms that prioritizing tokens via learned semantic importance, rather than rigid spatial anchors, enables the extraction of domain-invariant morphological features and effectively mitigates spatial overfitting.

\begin{figure}[t]
\centering
\includegraphics[width=0.98\linewidth]{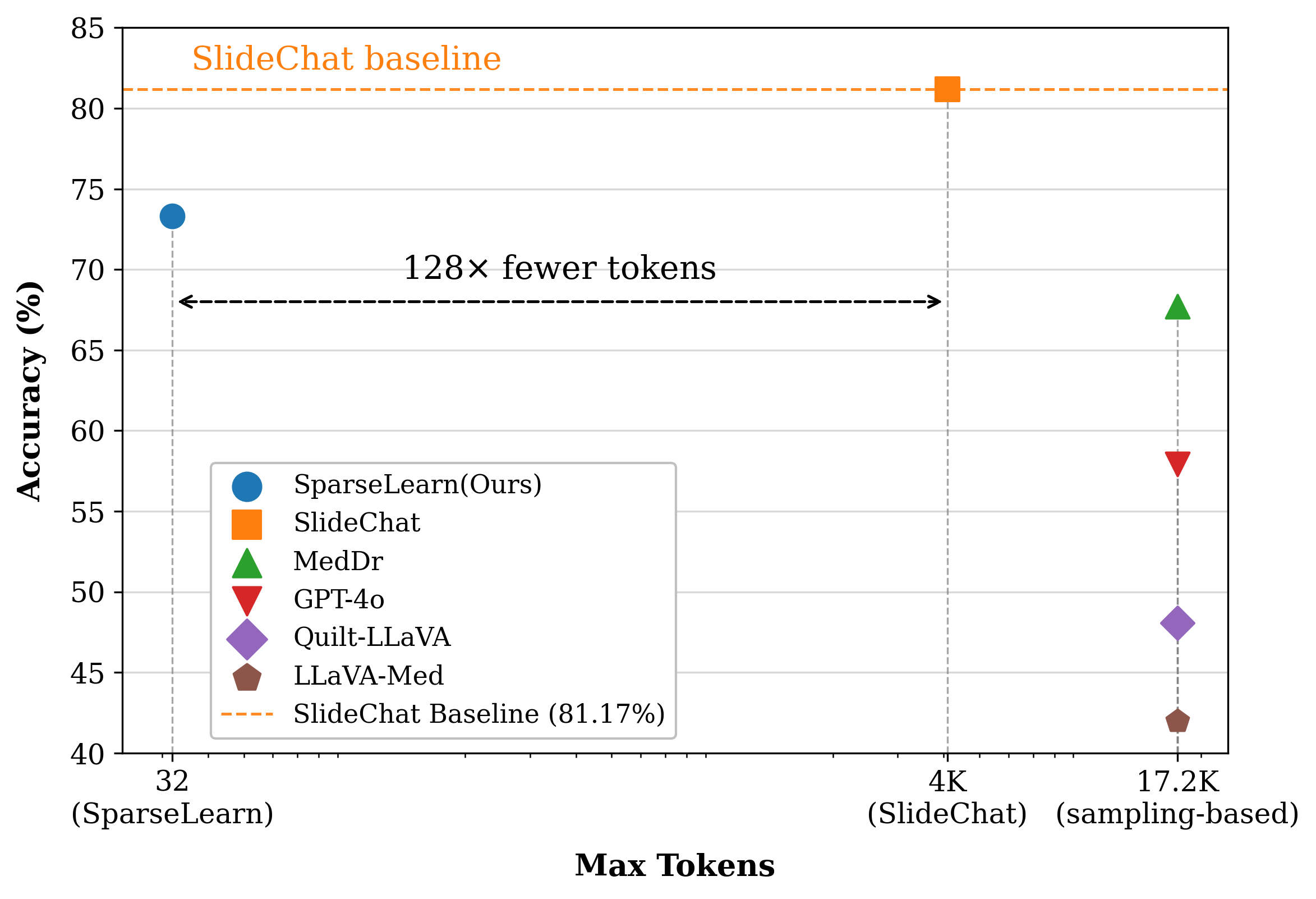}
\caption{\textbf{Accuracy and token efficiency.} Our framework achieves competitive diagnostic accuracy using only 32 tokens, which is 128 times fewer than the SlideChat \cite{chen2025slidechat} baseline. Furthermore, it significantly outperforms traditional patch-based VLM methods that demand over 17.2K tokens.}
\label{fig:token_efficiency}
\end{figure}

\begin{figure}[t]
\centering
\includegraphics[width=0.999\linewidth]{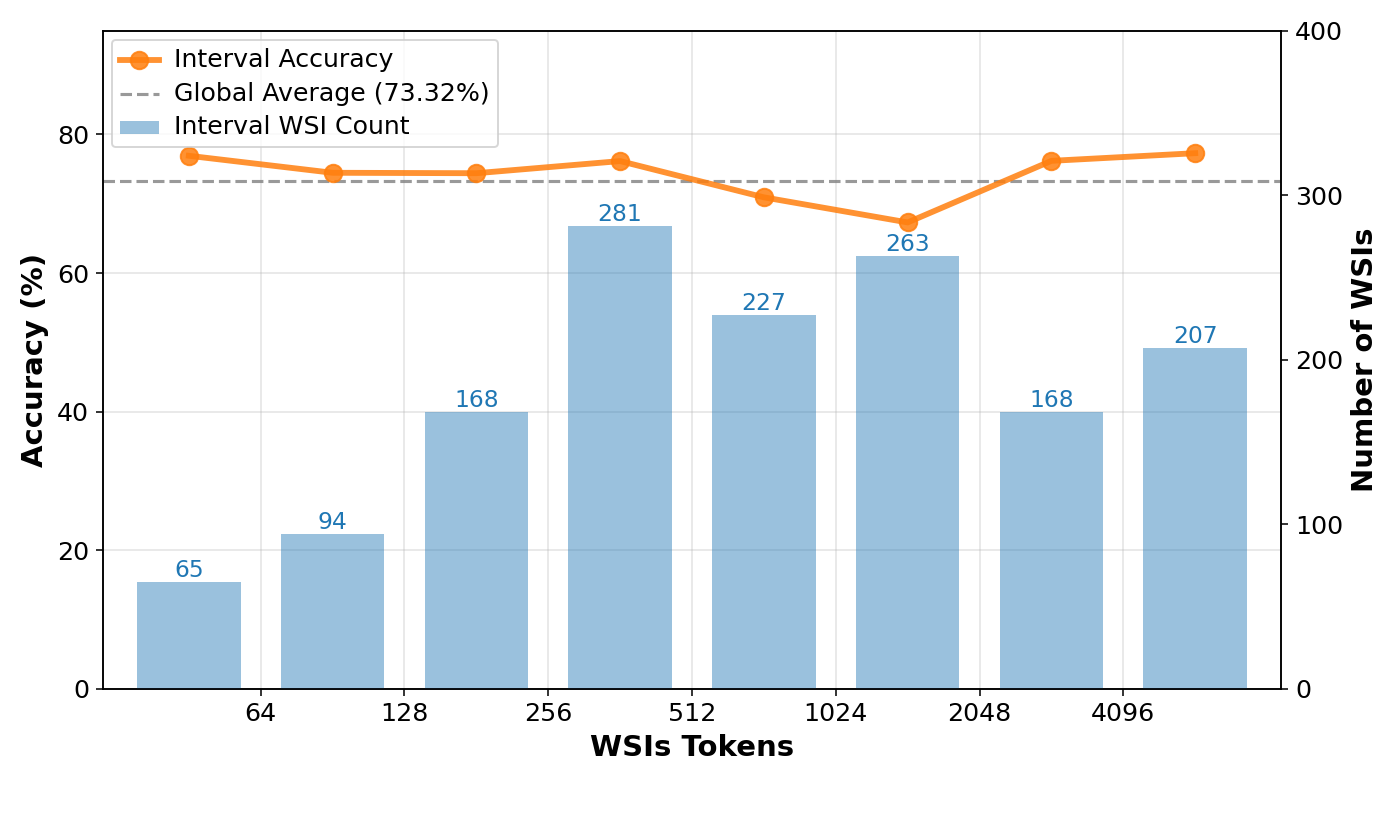}
\caption{\textbf{Performance stability across varying sequence lengths.} The interval accuracy (orange line) remains highly stable near the global average (73.32\%) despite exponential variations in WSI token counts (blue bars). This demonstrates that the model's reasoning capability is inherently insensitive to the input scale.}
\label{fig:performance_stability}
\end{figure}

\subsection{Efficiency Analysis}
\label{sec:efficiency_patho}

 \Cref{fig:token_efficiency} demonstrates the efficiency of the proposed method. While the baseline SlideChat operates with a maximum input capacity of 4,096 tokens, our framework achieves comparable performance with a fixed budget of only 32 tokens, representing a $128\times$ reduction in sequence length. Given that accuracy remains robust even for slides exceeding 4,096 patches, this 32-token representation establishes a dynamic limit for distilling gigapixel information. By fixing the sequence length at this compact scale, the quadratic computational overhead of the LLM reasoning stage becomes independent of the raw patch count $N$. This decoupling effectively mitigates the primary computational bottleneck, as the LLM no longer needs to process an expensive, variable-length visual prefix.
 
\Cref{fig:performance_stability} further illustrates our framework maintains consistent diagnostic fidelity across a wide spectrum of input scales. Although the raw sequence lengths range from 64 to over 10,000 patches, the interval accuracy remains stable near the global average of 73.32\%. This insensitivity confirms that the learnable selection mechanism effectively isolates an invariant semantic subspace, ensuring that reasoning is governed by intrinsic morphological features rather than the physical dimensions of the specimen.

\begin{figure}[t]
\centering
\includegraphics[width=1\linewidth]{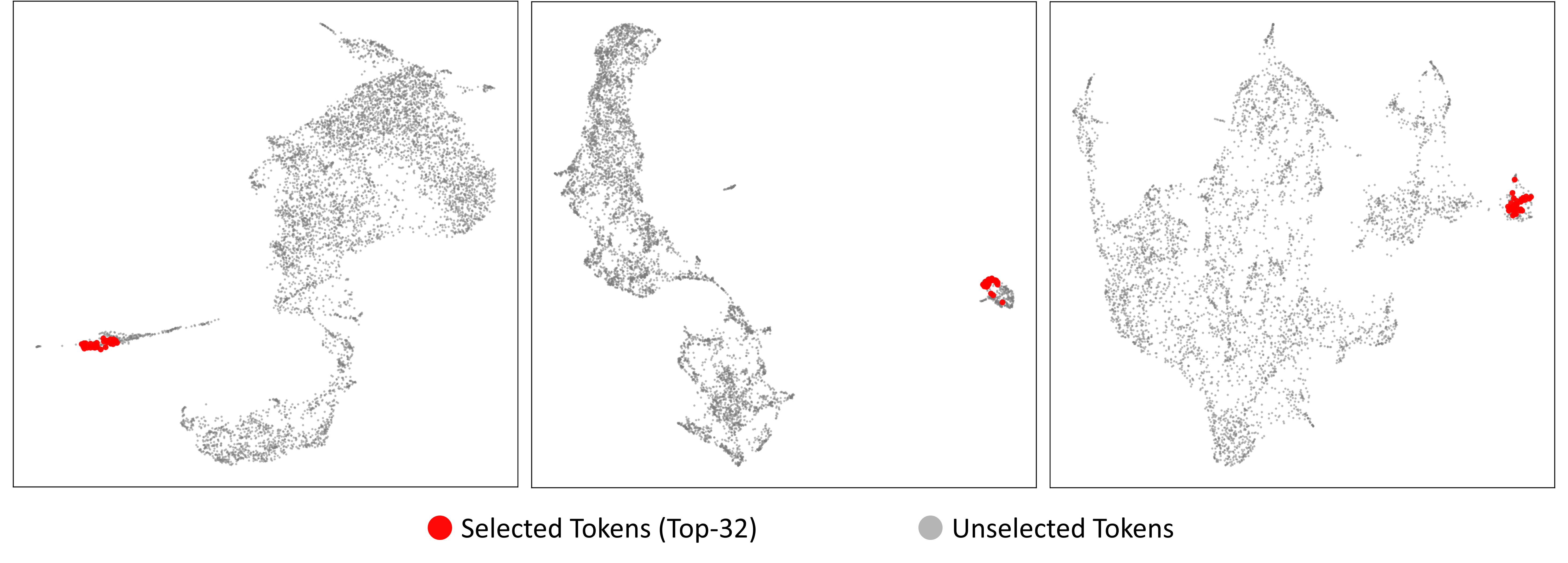}
\caption{UMAP visualizations of the latent feature space across three diverse WSIs. The selected tokens consistently cluster at topological extremities, demonstrating the model's precision in isolating sparse, discriminative morphological outliers without semantic dilution.}
\label{fig:UMAP}
\end{figure}

To qualitatively validate our selection mechanism, we project the high-dimensional visual features into a 2D topological space using UMAP (see~\Cref{fig:UMAP}). The vast majority of unselected tokens form dense central manifolds, corresponding to highly redundant stromal or healthy background tissues. In contrast, the 32 tokens retained by our approach consistently cluster at the extreme topological margins or isolated sub-clusters across diverse slides. This spatial divergence corroborates that our sparsity-constrained routing effectively pinpoints extremely sparse, highly discriminative morphological outliers.

\subsection{Ablation Studies}
\label{sec:ablation_patho}

\begin{table}[htbp]
\centering
\caption{\textbf{Ablation studies} results at $K=32$. Each row reports the accuracy. ``Base'' denotes the full SparseLearn configuration (VP noise + diagonal attention).}
\label{tab:ablation}
\large
\renewcommand{\arraystretch}{1.3} 
\setlength{\tabcolsep}{10pt} 
\begin{tabular*}{\linewidth}{@{\extracolsep{\fill}}llc}
\toprule
\textbf{Configuration} & \textbf{Modification} & \textbf{$K=32$} \\
\midrule
Base (VP noise + diagonal)$^{\dagger}$ & -- & 73.32\% \\
Scale gating & VP noise $\rightarrow$ scale & 70.40\% {\footnotesize (-2.92)} \\
Global attention & diagonal $\rightarrow$ global & 71.62\% {\footnotesize (-1.70)} \\
\bottomrule
\end{tabular*}
\end{table}


We evaluate the architectural necessity of the two key modules: the diagonal attention mechanism and the VP noise gating within the training phase.

\textbf{Diagonal vs. global attention.}  \Cref{tab:ablation} shows that replacing the diagonal attention in the denoiser with standard global self-attention causes a significant performance drop from 73.32\% to 71.62\%. This degradation confirms the semantic leakage hypothesis. Global attention allows low-importance tokens to incorporate information from high-score neighbors, effectively circumventing the intended sparsity bottleneck during training. Enforcing diagonal isolation is thus essential for ensuring that the information throughput of each token is strictly governed by its own importance score, thereby maintaining the integrity of the sparsity-constrained formulation.

\textbf{VP noise gating vs. scale gating.} Substituting our VP noise gating with a conventional scale-based gating mechanism (direct multiplication by $\alpha$) results in a decrease to 70.40\%. This performance gap highlights the sensitivity of the frozen LLM to distributional shifts in the visual prefix. Unlike scale gating, which shrinks feature magnitudes and causes latent collapse, VP noise gating preserves the macroscopic feature variance. This stochastic proxy more accurately simulates the information restriction of discrete token pruning, providing a more stable optimization target for the scorer and ensuring robust end-to-end continuous gradient flow.




\section{Conclusion}
\label{sec:conclusion}

In this work, we reformulate WSI token pruning as a learnable sparsity paradigm. By decoupling VP noise gating during training from deterministic Hard Top-$K$ token pruning at inference, the SparseLearn training module resolves the non-differentiability of discrete pruning without requiring architectural modifications or auxiliary losses. The overall framework demonstrates scale-invariance in gigapixel WSIs reasoning. By distilling the visual context into a strict 32-token sparsity budget and achieving up to a 128-fold reduction in sequence length, our approach yields 73.32\% overall accuracy on SlideBench (TCGA) and distinctly improves zero-shot generalization. This establishes an end-to-end paradigm that balances global receptive fields with optimal inference latency.

Regarding limitations, our framework relies on a localized Scorer, assuming diagnostic importance is fully decipherable at the individual patch level. For macroscopic morphological patterns or diffuse tissue margins, the lack of early cross-patch communication prior to the pruning bottleneck may omit broader relational context. Additionally, enforcing a static sparsity budget limits flexibility across diverse clinical applications. Future work will explore dynamic budget allocation and incorporate spatial-graph priors into the scoring mechanism to enable adaptive token pruning for pathology foundation models.


\bibliographystyle{unsrt}
\bibliography{main}

\end{document}